\documentclass[a4paper, 10pt, conference]{ieeeconf}      
\usepackage{graphicx}
\usepackage{booktabs}
\usepackage{multirow}
\usepackage{amsmath}
\usepackage{amssymb}
\usepackage{xcolor}

\IEEEoverridecommandlockouts                              
\title{\LARGE \bf
Enhancing Sign Language Teaching: A Mixed Reality Approach for Immersive Learning and Multi-Dimensional Feedback
}

\author{ Hongli Wen, Yang Xu, Lin Li,  Xudong Ru$^*$,  Xingce Wang,  Zhongke Wu
\thanks{All the author are coming from the School of Artificial Intelligence, Beijing Normal University, Beijing 100875, China.The email of Xudong Ru (corresponding author) is ruxudong@126.com }
}

\begin{document}

\maketitle
\thispagestyle{empty}
\pagestyle{empty}

\begin{abstract}


Traditional sign language teaching methods face challenges such as limited feedback and diverse learning scenarios. Although 2D resources lack reality sences, classroom teaching is constrained by a scarcity of teacher and methods based on VR and AR have relatively primitive interaction feedback mechanisms. This study proposes an innovative teaching model that uses real-time monocular vision and mixed reality technology. First, we introduce an improved hand-posture reconstruction method to achieve sign language semantic retention and real-time feedback. Second, a ternary system evaluation algorithm is proposed for a comprehensive assessment, maintaining good consistency with experts in sign language. Furthermore, we use mixed reality technology to construct a scenario-based 3D sign language classroom and explore the user experience of scenario teaching. Overall, this paper presents a novel teaching method that provides an immersive learning experience, advanced posture reconstruction, and precise feedback, achieving positive feedback on user experience and learning effectiveness.

\end{abstract}

\section{INTRODUCTION}

Sign language teaching research faces numerous challenges\cite{ref1}, such as the convenience of 2D books and videos for self-study, but the lack of real-time feedback makes it difficult for learners to apply in practice; sign language action assessment is limited to bottom-up end-to-end recognition \cite{ref4,ref5,ref6}, lacking prior knowledge, leading to structural complexity and overfitting; although virtual reality and augmented reality technologies have opened new paths for sign language teaching \cite{ref2,ref3}, the immature feedback mechanisms of these platforms, and the uniformity of learning scenarios and low accuracy of action assessment, remain significant barriers to improving learning outcomes.

Hand reconstruction, as the primary link in three-dimensional sign language teaching, affects its feasibility in educational settings due to issues such as reconstruction effects and costs \cite{ref25,ref26,ref27}. EasyMocap, proposed by Shuai et al. \cite{ref11}, effectively recovers hand positions from RGB images, but its multi-view design makes it costly; FrankMocap, proposed by Yu et al. \cite{ref12}, can capture body and hand postures in natural environments, but has poor real-time performance (9.5 fps). Effective and reasonable sign language action assessment is key to teaching. Traditional distance measurement methods used in previous studies on action similarity calculation often fail to meet accuracy needs \cite{ref16}, such as the Euclidean and Manhattan distances used by Morais et al. \cite{ref15}. Shen et al. \cite{ref18} introduced a method for calculating quaternion distances, using cascaded quaternions to represent joint directions, to meet more precise and computationally efficient needs. Feedback-driven interactive learning can enhance the efficiency of sign language teaching \cite{ref20}.  Zhang et al. \cite{ref32} developed a sign language teaching system on smartphones to score learners' sign language,however,they failed to pinpoint the inaccurate parts of sign language.Suman Deb et al. \cite{ref29} launched an MR mobile application that displays 3D sign language animations by scanning pre-marked word cards, however, their method only includes hand models, lacks precise and diverse feedback, making it difficult for users to understand intentions.

To address these challenges, we propose an innovative sign language teaching model using technological means to overcome these limitations. Firstly, improvements have been made in single-view gesture reconstruction technology \cite{ref7}, which can extract hand movements from frame sequences in real-time and provide immediate feedback to learners. Next, we introduce a high-confidence tripartite system for sign language assessment based on action similarity evaluation, abandoning traditional end-to-end gesture recognition methods that lack explanatory power \cite{ref8} or direct regression scoring methods \cite{ref9}. This algorithm assesses actions from three perspectives to achieve high consistency with expert evaluations. Finally, to enhance learning outcomes, this study explores the concept of building a scene-based 3D classroom using mixed reality technology, enhancing learners' memory and understanding through an immersive learning environment. Through these innovative methods, this study aims to provide a more effective and efficient new model for sign language learning, with the following main contributions:

\begin{itemize}

\setlength{\leftmargin}{0pt} 
    \item We constructed an immersive mixed-reality 3D sign language teaching pipeline across time and space. It separates teaching from learning scenes, collects teaching materials from multiple channels, and reconstructs teaching segments in mixed-reality environments.
   
    \item Introduced a posture reconstruction scheme with hands based on a high-precision human body mesh reconstruction method using monocular vision \cite{ref7}. Based on the component-aware transformer, it encodes the whole body and decodes locally, recovering poses in various temporal and spatial conditions in real time, thus constructing a sign language teaching database.
    
    \item A two-stage action assessment method is proposed, where the first stage involves gesture reconstruction and the second stage scores based on comparing the similarity of sign language actions in 3D space, providing precise feedback on the continuity, similarity, and accuracy of hand movements.
\end{itemize}

\section{SYSTEM DESIGN}
\subsection{Task Description}

\begin{figure*}[]
\centering 
\includegraphics[width=0.9\textwidth]{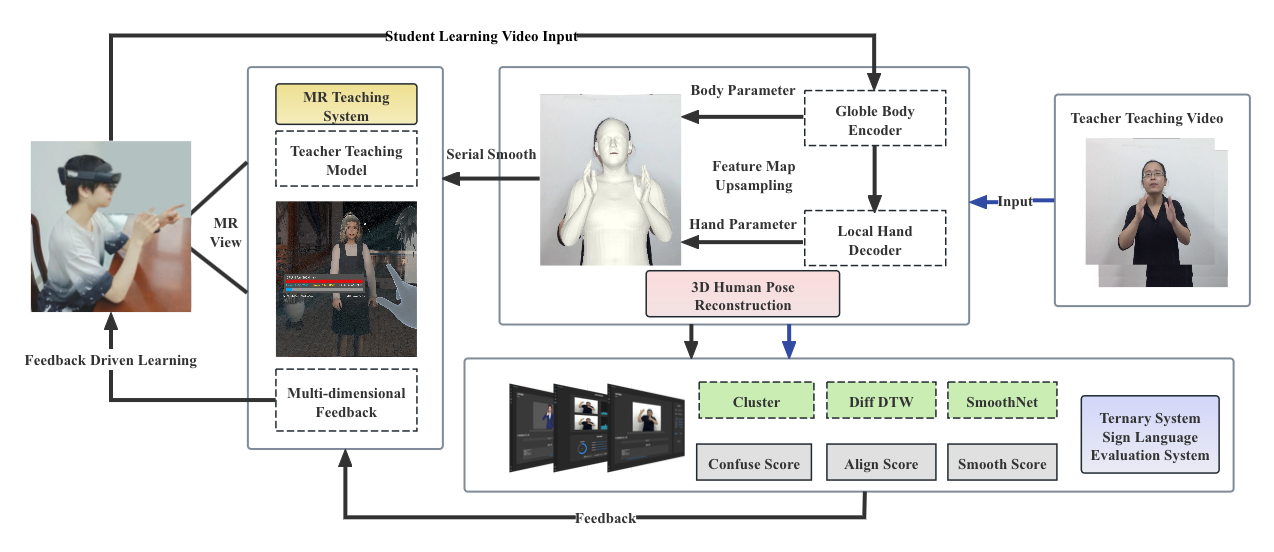} 
\caption{Sign language teaching system architecture} 
\label{Fig.1} 
\end{figure*}

Based on MR, we propose a three-dimensional sign language teaching system that utilizes a 3D reconstruction method to accurately capture the motion characteristics of teaching instructors from videos to generate 3D teaching data. Virtual scene construction and teaching take place on MR devices (HoloLens2). Through timely feedback, learners can obtain scores and modify their actions based on instructions.
Our system shown in Fig.\ref{Fig.1} consists of three parts with the tasks  as follows.

First, the 3D sign language reconstruction and presentation part extracts independent spatiotemporal data from real sign language movements, smoothing and quantitatively modeling them while preserving their semantic basis. We refine a one-stage 3D human body reconstruction method, focusing on sign language, optimizing for real-time and accuracy.When performing in students,it  works also as first stage of action quality assessing. 
Second, the 3D sign language teaching part utilizes standard sign language data from professional instructors, generating teaching scenes where virtual avatars teaching. Pose redirection drives these avatars, and students interact with them on mixed reality terminals. 
Finally, the sign language teaching feedback system accepts standard sign language input, providing feedback through confusion scores, fluency scores, and completeness evaluation, giving students meaningful feedback information.

\subsection{3D Pose Reconstruction for Enhanced Accuracy and Real-Time Sign Language Teaching}

To construct a corresponding skeleton from RGB monocular image sequences for motion feedback\cite{ref17}, We introduce a one-stage pose reconstruction algorithm that is suitable for estimating hand poses based on Vit-Pose using a module-perceiving transformer model.This approach utilizes a module-perceiving transformer model that estimates body posture and gestures simultaneously. It includes a global encoder and a local decoder to enhance the accuracy of hand keypoints and enable real-time extraction of sign language motion data.

To improve the computational efficiency and reduce the computational pressure, we segment the input image into fixed-size image blocks, where $M$ is the size of the image block, $\mathbf{P}$ represents the image blocks, and $H$ and $W$ are the height and width of the input image, respectively. This allowed us to process different parts of the image separately,
\begin{equation}
\mathbf{P}\in\mathbb{R}^{\frac{HW}{M^{2}}\times(M^{2}\times3)}.
\end{equation}

Considering that the resolution of the hand image blocks is low, direct pose estimation may lead to insufficient accuracy. Therefore, a feature-level upsampling and cropping strategy was employed in the local decoder. Specifically, the feature-token sequence $\mathbf{T_{f}}^{\prime}$ obtained from the global encoder is reshaped into a feature map, which is upsampled through deconvolutional layers to obtain multiple higher-resolution features $\mathbf{T}_{hr}$, thereby providing more detailed hand features. Additionally, the decoder uses keypoint queries $\mathbf{Q}$ extract features from the elements of multi-scale features $\mathbf{V}$ around the position of keypoints $p_q$:
\begin{equation}
CA(\mathbf{Q},\mathbf{V},p_q)=\sum_{l=1}^L\sum_{k=1}^KA_{lqk}W\mathbf{V}l(\phi_l(p_q)+\Delta p{lqk})
\end{equation}
where $l$ and $k$ represent the feature level and keypoint, respectively. $CA$ represents the Cross-Attention function, which selectively emphasizes and integrates features from different layers and keypoints to improve the accuracy and detail of the pose estimation. $A$ and $W$ are the attention weight and learnable parameters, $L$ refers to the number of layers in the multi-level feature maps, and $K$ is the number of keypoints used to precisely capture features.respectively. $\phi(\cdot)$ and $\Delta p$ represent the position adjustments and offsets, respectively. This attention mechanism focuses on the local features around hand keypoints, increasing accuracy while reducing unnecessary computational overhead.
Our method of 3D pose reconstruction based on monocular vision provides a real-time action generation solution, and offers the ability to process and reuse data across time and space. This provides ample momentum for sign language teaching.


\subsection{3D Sign Language Teaching}
\subsubsection{Teaching Pipeline}
We utilized pose recognition algorithms to fully reconstruct the 3D body and posture of the teacher from 2D videos, capturing the full-body posture, facial expressions, and hand movements. These movements were reconstructed on a standard SMPLX model and smoothed using SmoothNet, thereby enhancing their expressiveness for teaching purposes. Standard teacher movements were stored in a database for motion alignment and feedback.By building this teaching pipeline, students can independently learn by  mimicking the actions of the 3D sign language teacher and receive  immediate feedback, enhancing learning efficiency.

\subsubsection{Interface Interaction and Design of the Teaching System}
\begin{figure}
\centering 
\includegraphics[width=0.45\textwidth]{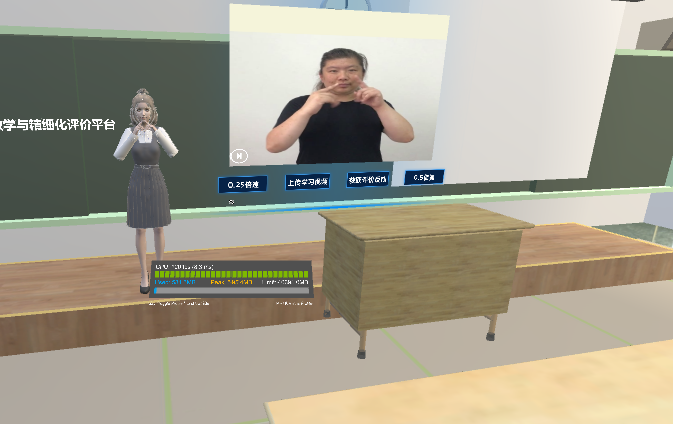} 
\caption{Interactive interface of the teaching system} 
\label{Fig.4} 
\end{figure}

Within the framework of the post-WIMP paradigm, we developed an MR-supported 3D Chinese sign language teaching and evaluation system. The system’s user interface comprises two modules: a learning module and a feedback module.In the learning module, users select a specific vocabulary or sentence for practice, receiving 3D sign language teaching from a virtual instructor tailored to their selections.Upon switching to the feedback module, the system captures the user’s sign language movements via a camera and transmits the video to the server for analysis. The system then provides comprehensive feedback on posture accuracy, movement coherence, and confidence level.Through the interaction of the learning and feedback modules, students  can not only view the details of sign language actions from multiple  angles, but also adjust their gestures in real time based on feedback,  accelerating the learning process.The interactive interface of the teaching system is shown in Figure \ref{Fig.4}.

\subsection{Learning with Feedbacks}
\begin{figure}[htbp] 
\centering 
\includegraphics[width=0.45\textwidth]{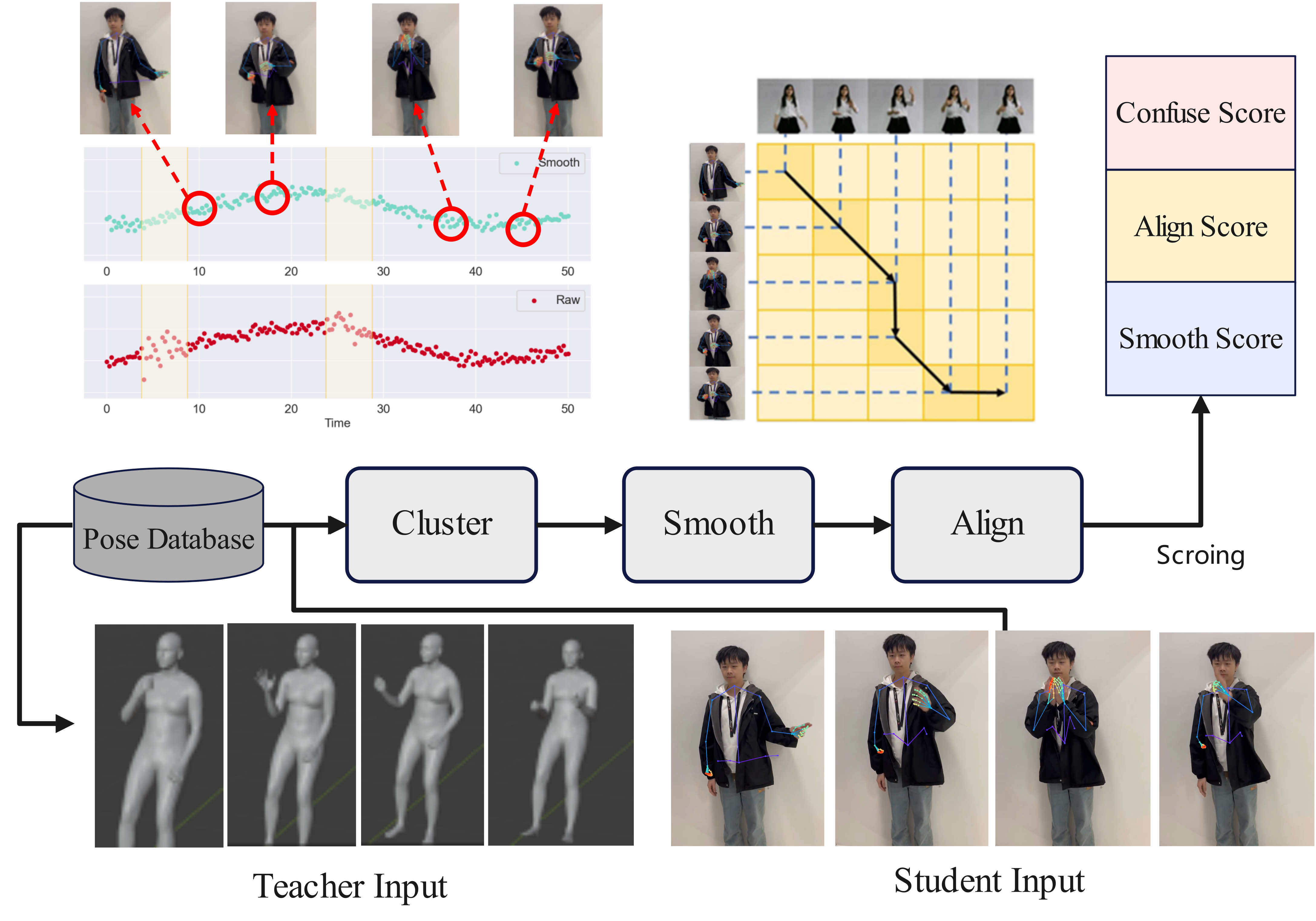} 
\caption{Sign language action evaluation algorithm based on ternary system} 
\label{Fig.7} 
\end{figure}

Our sign language teaching systems can provide real-time feedback on student learning outcomes. We adopted a two-stage top-down approach, where the first stage involved posture reconstruction, as previously described, and the second stage compared the spatiotemporal features of movements.

\subsubsection{Quaternion-based Sign Language Action Embedding}
The human body model comprises \(N\) joints, the rotation of the \(i\)-th joint can be represented by a quaternion \(q_i\). The position transformation of the joint relative to its parent joint was calculated using (3),

\begin{equation}
v'_i = q_i v_i q_i^{-1} 
\end{equation}

\noindent where \(v_i\) denotes the initial position vector of the \(i\)-th joint in the parent joint coordinate system, \(v'_i\) denotes the rotated position vector, \(q_i\) represents the quaternion rotation of the \(i\)-th joint.

For entire body, this transformation must be applied recursively, starting from the root joint and calculating the global positions of all joints through local rotations:

\begin{equation}
v'_{\mathrm{global},i} = \prod_{k=1}^{i} q_k v_i \left( \prod_{k=1}^{i} q_k \right)^{-1}
\end{equation}

\noindent where \(\prod_{k=1}^{i} q_k\) represents the cumulative product of all the rotation quaternions along the path from the root joint to the \(i\)-th joint.

Although it is possible to compute the distance between action sequences to assess differences, the dimensions are not independent. One quaternion depends on another, and differences in the parent joints directly affect the differences in the child joints. Therefore, we used a weighted embedding method to correct this and obtain a low-dimensional representation of the differences, enabling the direct calculation of the distance between each frame's movements.

The student action sequence counted in frame \(t\) is defined as \(V_{\text{stu}}(t)=v'^{\text{stu}}_{\mathrm{global}}(t)\). According to forward kinematics, this representation unfolds into a high-dimensional tensor at each specific time point \(t'\) as \([p^{(1)},p^{(2)},...,p^{(N)}] \in \mathbb{R}^{4 \times N}\), where each joint dimension \(p^{(i)}\) is recursively computed for the parent joint. The teacher sequence was expressed as \(V_{\text{tea}}(t)\). We used quaternion logarithmic differences, where \(q^*\) represents the conjugate of \(q\):
\begin{equation}
    d_i(t_1,t_2)=\log(q_{\text{tea}}^i(t_1) \cdot (q_{\text{stu}}^i(t_2))^*)
\end{equation}
     
\noindent then we calculate the embedding representation of each joint \(D_i(t_1,t_2)\), where \( D_0(t_1,t_2)=\mathbf{0} \):
{\small
\begin{equation}
     D_i(t_1,t_2)  =  \mathbf{W}^i\frac{ d_i(t_1,t_2) }{m_1D_{i-1}(t_1,t_2) }+m_2D_{i-1}(t_1,t_2) 
\end{equation}
}

\noindent where $\mathbf{W}$ is a learnable weight matrix that performs affine transformation.

\subsubsection{Confusion Feedback}

First, we evaluated the discriminability of a student's sign language, which is referred to as action confusion. According to the  distance  above, a sign language segment is regularized to a lower dimension \(n\), and in this low-dimensional space, the teacher's sign language actions are clustered.

Using a fixed clustering method, student action data \(V^s\) are assigned to the nearest cluster \(K_s\). We then use the distance \(d(P^s, P^t_i)\) between actions to estimate the probability \(\delta(P^s, P^t_i)\) that a student action belongs to a teacher action \(P^t_i\) to obtain the probability distribution of student actions in the teacher action space \(\delta(P_k)\) shown in (7).
\begin{equation}
    \delta(P_k) = \mathrm{softmax}(d(P^s, P^t_i))
\end{equation}

The concentration and dispersion of this probability distribution serve as an evaluation of the action confusion \(\mathrm{C}\). This metric guides the students in analogizing and distinguishing similar actions.

\subsubsection{Smoothness Evaluation}

As the smoothness of actions is highly indicative of learning progress in sign language, reflecting the coupling between body movements and muscle memory, we introduced SmoothNet\cite{ref35} to smooth the original student action sequence \(P^s\) in the same way as teacher action extraction, estimating an idealized action sequence \(P_{\mathrm{smooth}}^s\) at the ideal time scale.

We then compared it with the original sequence \(P_{\text{smooth}}\), calculated the distance \(d_s\), estimated the smoothness \(l_{\mathsf{smooth}}\) of the student action, and obtained a smoothness score \(\mathrm{S}\). This metric evaluates the fluency and coherence of student actions while providing an ideal smooth sequence as a reference for error correction.

\subsubsection{Alignment Feedback}

We have the same form of teacher action sequence \(P^{t}_{\mathrm{smooth}}\) and student  sequence \(P^{s}_{\mathrm{smooth}}\). Due to differences in proficiency, intensity, and coherence between individuals, these two sequences typically have different lengths in the time dimension. Conventional distances such as Euclidean distance, cannot provide accurate evaluations. Therefore, we adopted derivative-based Dynamic Time Warping (DTW)\cite{ref28,ref28} to detect the gradient of joint movements in the smoothed sequence.

Interpolating quaternions between discrete frames can calculate the size of gradients, which serve as a measure of motion trends and can, to some extent, eliminate differential features by focusing on the characteristics of the sign language itself. The adaptive joint-weighted distance \(D(t)\) of each action sequence is calculated in the time dimension by considering the gradient \(\nabla_{\phi_i}(t)\) of the motion as (8).
\begin{equation}
    D = \text{DTW}(\nabla_{\phi_{\text{stu}}}(t),\nabla_{\phi_{\text{tea}}}(t))
\end{equation}

\section{EXPERIMENT}

The experiments are divided into four parts to verify the effectiveness of the proposed system. First, the real-time performance and accuracy of the sign language reconstruction algorithm were evaluated; second, the credibility and effectiveness of the action feedback algorithm were evaluated using the collected data; and finally, the overall effect of the system was evaluated through user experience testing.

\subsection{Construction of a Multimodal Synchronous Chinese Sign Language Dataset}

We constructed a multimodal synchronous Chinese Sign Language dataset for these experiments below. The dataset was based on numerous monocular sign language teaching videos provided by the Chinese Academy of Sciences. The dataset includes teaching sentences and vocabulary items segmented for modular and lexical-level sign language teaching. Each sign language vocabulary item also includes 3D sign language action data generated by our algorithm.

\begin{table}[h]
\centering
\caption{Dataset content and quantity}
\resizebox{0.45\textwidth}{!}{%
\begin{tabular}{lc}
\hline
Multimodal Synchronous Dataset Content Type & Quantity (items)\\
\hline
Sentence videos (.avi) & 1280 \\
Sentence texts (.txt) & 1280 \\
Vocabulary videos (.avi) & 568 \\
Vocabulary texts (.txt) & 568 \\
Motion data files (.bvh) & 568 \\
\hline
\end{tabular}}
\end{table}

 Motion files of sign language vocabulary were used to generate virtual teaching segments for teachers. loaded into mixed reality teaching devices and provided students with Chinese Sign Language in another teaching environment, realizing a spatiotemporal heterogeneous teaching mode.

\subsection{Verification of the Accuracy and Real-Time Performance of the Posture Reconstruction Method}
We utilized 3D reconstructed movements, prioritizing accuracy for teacher movements and real-time performance for student movements.

Nine sign language experts, each with over five years of teaching experience, evaluated 65 categories of sign language movements recognized by different algorithms, focusing on semantic accuracy and fluency, with scores up to 10 points.The results are listed in the table~\ref{tab:1}:

\begin{table}[h]
\centering
\caption{Evaluation results of different algorithms}
\label{tab:1}
\begin{tabular}{lccc}
\hline
Method & Semantic Accuracy (out of 10) & Fluency (out of 10)\\
\hline
Frankmocap         & 8.92         &7.90\\
Hand4Whole         & 8.84         &8.15\\
SMPLer-X         & 9.76         &8.58\\
Easymocap         & 8.98         &8.54\\
Expose         & 9.38         &8.49\\
Ours     & \textbf{9.92}         &\textbf{8.78}\\
\hline
\end{tabular}
\end{table}

The results demonstrate our algorithm's advantage in accurately capturing sign language semantics and simulating natural movements, reflected in a low average keypoint error of 35.7mm on the hand. This precision makes it highly suitable for sign language demonstration teaching.

About the interface speed (FPS), we found our algorithm performed well in terms of FPS (38.6), indicating high real-time performance for a better interactive teaching experience.

\subsection{Verification of the Credibility and Accuracy of the Sign Language Evaluation Algorithm}
\textbf{Experiment Design.}
To verify the credibility and accuracy of the scoring function of the platform, we selected 15 common sign language phrases. Each sign language phrase was recorded by six participants with a certain level of sign language learning experience, resulting in 90 videos (15 sign language phrases, $\times$ 6 videos). We invited five sign language experts to rate and rank 15 sets of videos on our platform. The rating criteria included the coherence, similarity, and accuracy of sign language movements.
We use the Spearman Rank correlation $\rho = \mathrm{cov}(R_{p}, R_{g}) / \sigma_{R_{p}} \sigma_{R_{g}}$ as the evaluation metric, where $R_{p}$ is the platform's predicted score rank and $R_{g}$ is the expert rating rank. A higher Spearman correlation indicates that our platform has better rank prediction ability.


\textbf{Results.} The average Spearman correlation coefficient between the expert and platform ratings for the 15 test sets was 0.86. This indicates that the rating algorithm of the platform is highly consistent with the expert ratings of sign language movements. The evaluations provided by sign language experts on the coherence, similarity, and accuracy of the movements were highly consistent with the results calculated by our evaluation algorithm. This demonstrates that the feedback provided by the platform to students was highly credible and accurate. This can, to a certain extent, reduce the cost of sign language teacher ratings and guidance and improve teaching efficiency.

\subsection{Verification of the Effectiveness of the 3D Sign Language Teaching System}

\textbf{Experiment Design.}
This experiment aimed to validate the effectiveness of a comprehensive system for Chinese Sign Language teaching and evaluation based on mixed reality (MR) technology. To increase the representativeness and general applicability of the experimental results, we invited individuals of different ages and genders with no prior experience using MR devices to participate in the experiment.We divided the participants into three groups: A, B, and C. Group A participants learned by watching 2D sign language teaching videos. Group B used our teaching system with evaluation feedback for learning, and Group C used a teaching system without evaluation feedback. The differences in the learning effects before and after the study were evaluated using our system for all participants.

\textbf{Results.}
We used t-tests to analyze the age group and gender data for the three learning methods, and the results are shown in Table~\ref{tab:2}.The results indicate that there was no significant difference in learning effects among different age groups and genders for the three learning methods. This suggests that age and sex were not key factors affecting the improvement in learning effects in this experiment.

\begin{table}[htbp]
\centering
\caption{t test results of age group and sex group}
\label{tab:2}
\begin{tabular}{ccccc}
\toprule
\multicolumn{2}{c}{}                              & Group A         & Group B         & Group C         \\
\midrule
\multirow{2}{*}{Age group} & t Stat & -2.25      & -1.76 & -1.95  \\
                        & P-value     & 0.13 & 0.16 & 0.15 \\
\midrule
\multirow{2}{*}{Sex group} & t Stat & -1.19 & -1.10 & -1.36 \\
                        & P-value     & 0.22 & 0.23 & 0.20 \\
\bottomrule
\end{tabular}
\end{table}

\begin{figure}[htbp] 
\centering 
\includegraphics[width=0.45\textwidth]{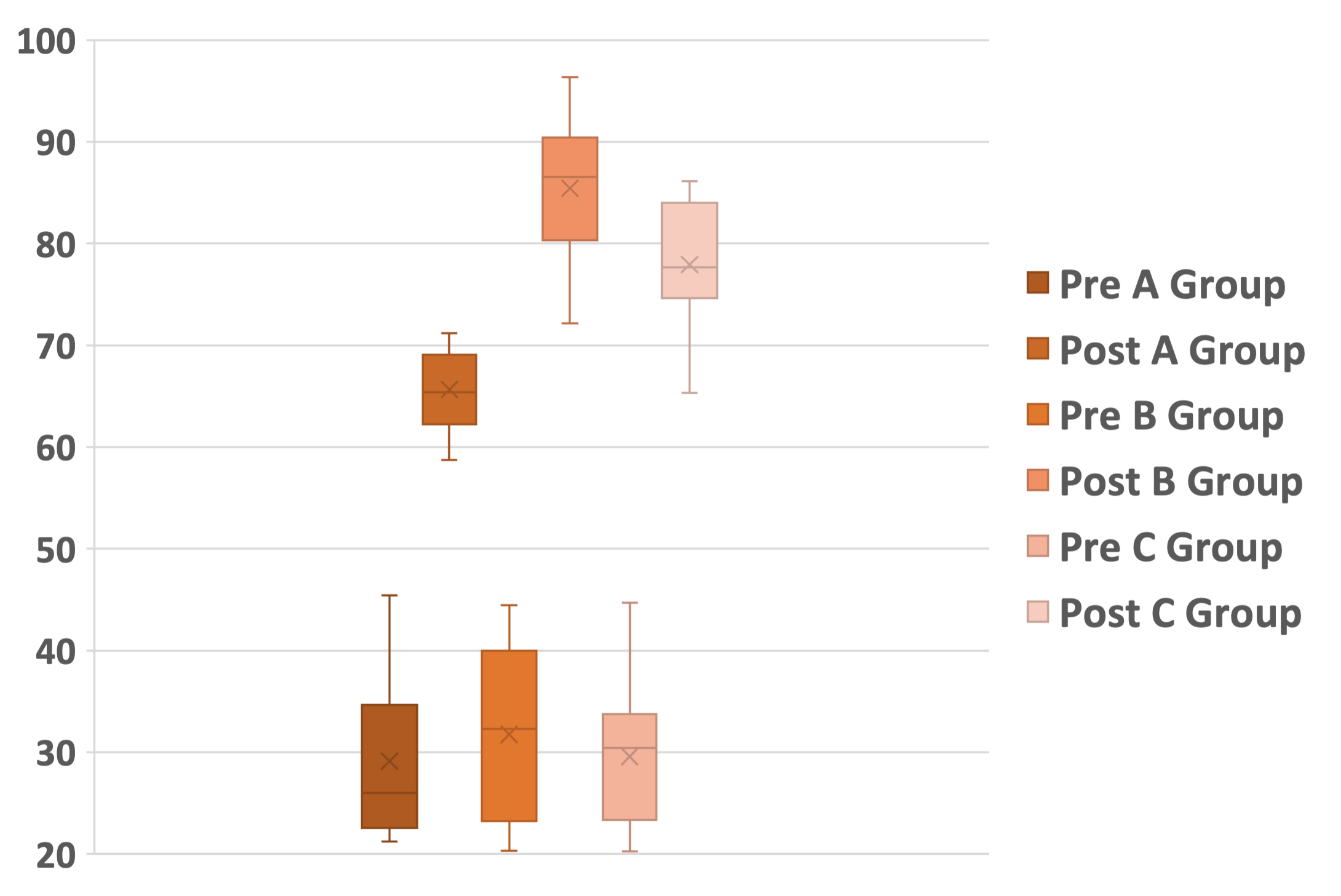} 
\caption{Pre-test and post-test box diagrams} 
\label{Fig.9} 
\end{figure}

We also analyzed the scores of the participants in Groups A, B, and C and found significant differences in the pretest and posttest scores among the three groups using a t-test (p\textless 0.05), as shown in Table ~\ref{tab:3}.This result indicates that all three learning methods promote effective learning of sign language to some extent.The mean, median, maximum, minimum, and quartiles of the scores before and after learning for Groups A, B, and C are shown in Figure ~\ref{Fig.9}.

\begin{table}[htbp]
\centering
\caption{Pre-test and post-test scores t test results}
\label{tab:3}
\begin{tabular}{cccc}
\toprule
       & Group A & Group B & Group C \\
\midrule
t Stat & -16.54  & -21.19  & -22.36  \\
P Value    & 2.03E-09 & 5.79E-14 & 1.18E-12 \\
\bottomrule
\end{tabular}
\end{table}

Groups A, B, and C participants experienced three different learning methods for sign language, and their scores increased by an average of 36.54, 53.67, and 48.36, respectively. Group B participants used a teaching system with evaluation feedback, allowing learners to quickly correct mistakes based on scoring results, resulting in the most significant improvement in learning effectiveness.

\subsection{Subjective Satisfaction Test}
To further test the effectiveness of our system, after learning through the 3D sign language teaching system, we surveyed 60 users on their subjective satisfaction with four sign language teaching methods: 2D sign language, traditional offline teaching by teachers, 3D sign language teaching without scenes, and 3D sign language teaching with scenes.We collected the user choices for the four teaching methods based on five indicators: motivation, pleasure, efficiency, satisfaction, learnability, and usability. The statistical analysis results were as follows:

\begin{figure}[htbp] 
	\centering
 \includegraphics[width=0.45\textwidth]
 {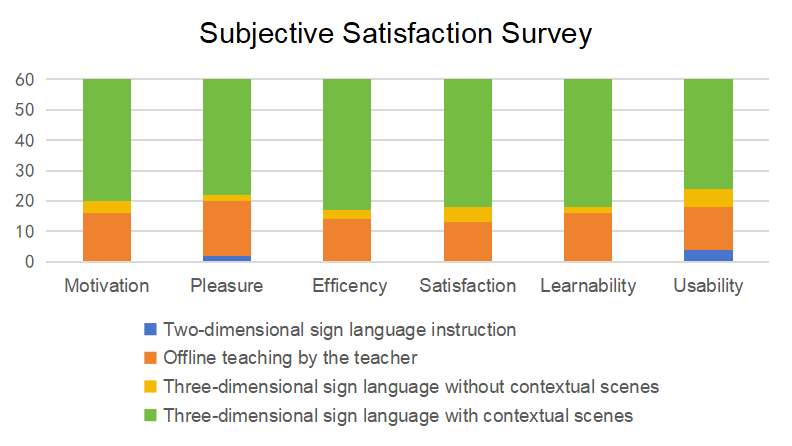} 
	\caption{User subjective satisfaction with different sign language learning methods}
	\label{fig:9}
\end{figure}

More than 60\% of the users believe that our 3D sign language teaching system is of immense help in terms of motivation, satisfaction and so on. This indicates that our system has a high level of recognition and practicality among users, providing strong support for the application and promotion of our system in sign language teaching.

\section{Conclusion and Future Work}

This study proposes a mixed-reality-based Chinese Sign Language teaching and feedback system.Our system can robustly assess the sign language skills of users and provide accurate and objective evaluations. Users of our system outperformed students taught using traditional methods.In our future work, we plan to introduce intelligent algorithms to provide users with personalized and customized feedback and suggestions. In addition, the proposed system framework can be applied to research in other fields such as assisting studies involving practice exercises for patients with brain injuries.






\end{document}